\documentclass[]{fairmeta}

\usepackage{wrapfig}
\usepackage{tabularx}
\usepackage{amsmath,amssymb,amsfonts}
\usepackage{algorithmic}
\usepackage{graphicx}
\usepackage{textcomp}
\usepackage{bbding}
\usepackage{pifont}
\usepackage{wasysym}
\usepackage{amssymb}
\usepackage{xcolor}
\usepackage{tcolorbox}
\usepackage{booktabs}
\usepackage{tabularx}
\usepackage{ragged2e}
\usepackage{amsmath}
\usepackage{multirow}
\def\BibTeX{{\rm B\kern-.05em{\sc i\kern-.025em b}\kern-.08em
    T\kern-.1667em\lower.7ex\hbox{E}\kern-.125emX}}

\newlength\savewidth

\newcolumntype{x}[1]{>{\centering\arraybackslash}p{#1pt}}
\newcolumntype{y}[1]{>{\raggedright\arraybackslash}p{#1pt}}
\newcolumntype{z}[1]{>{\raggedleft\arraybackslash}p{#1pt}}

\renewcommand{\paragraph}[1]{\vspace{1.25mm}\noindent\textbf{#1}}

\usepackage{algorithm}
\usepackage{listings}

\definecolor{codeblue}{rgb}{0.25, 0.5, 0.5}
\definecolor{codekw}{rgb}{0.35, 0.35, 0.75}
\lstdefinestyle{Pytorch}{
    language = Python,
    backgroundcolor = \color{white},
    basicstyle = \fontsize{9pt}{8pt}\selectfont\ttfamily\bfseries,
    columns = fullflexible,
    aboveskip=1pt,
    belowskip=1pt,
    breaklines = true,
    captionpos = b,
    commentstyle = \color{codeblue},
    keywordstyle = \color{codekw},
}

\definecolor{green}{HTML}{009000}
\definecolor{red}{HTML}{ea4335}

\title{Synthesizing Public Opinions with LLMs: Role Creation, Impacts, and the Future to eDemorcacy}

\author[1\dagger]{Rabimba Karanjai}
\author[1]{Boris Shor}
\author[1]{Amanda Austin}
\author[3]{Ryan Kennedy}
\author[1]{Yang Lu}
\author[2]{Lei Xu}
\author[1]{Weidong Shi}

\affiliation[1]{University Of Houston}
\affiliation[2]{Kent State University}
\affiliation[3]{Ohio State University}
\contribution[\dagger]{rabimba@cs.uh.edu}

\abstract{
This paper investigates the use of Large Language Models (LLMs) to synthesize public opinion data, addressing challenges in traditional survey methods like declining response rates and non-response bias. We introduce a novel technique: role creation based on knowledge injection, a form of in-context learning that leverages RAG and specified personality profiles from the HEXACO model and demographic information, and uses that for dynamically generated prompts. This method allows LLMs to simulate diverse opinions more accurately than existing prompt engineering approaches. We compare our results with pre-trained models with standard few-shot prompts. Experiments using questions from the Cooperative Election Study (CES) demonstrate that our role-creation approach significantly improves the alignment of LLM-generated opinions with real-world human survey responses, increasing answer adherence. In addition, we discuss challenges, limitations and future research directions.
}

\date{\today} 
\begin{document}

\maketitle

\section{Introduction}
Public opinion research, a cornerstone of democratic societies, has faced significant challenges in recent decades \cite{berinsky2013silent}. One of the most pressing issues is the growing difficulty in obtaining survey data that accurately represents a population \cite{plewes2013nonresponse}. A key problem lies in the steep decline in response rates for traditional survey methods such as telephone surveys. This trend complicates efforts to collect data with adequate sample sizes, particularly when analyzing specific subgroups \cite{kennedy2019response}. The drop in response rates exacerbates non-response bias, which becomes significantly pronounced when response patterns vary systematically across demographic or political subpopulations, such as those defined by age, race, or political affiliation \cite{simmons2023large}. This issue is further compounded when the bias aligns with unobservable yet critical characteristics, like political ideologies or voting behaviors \cite{groves2008impact}. 

Even surveys with large overall sample sizes can struggle with insufficient data points \cite{wang2015forecasting}. This challenge, often called the "curse of dimensionality," undermines the validity of inferences about subpopulations and poses significant hurdles for studying political behavior and public opinions \cite{bellman1957dynamic, ornstein2020stacked}.


In response to these obstacles, social scientists have embraced innovative approaches such as multilevel regression with poststratification (MRP) \cite{gelman1997poststratification, park2006state}. MRP and its variants have emerged as essential tools for estimating opinions within subgroups, especially in hierarchical or multilevel data structures such as regional and demographic categories. These methods improve accuracy by leveraging information from larger groups to generate more reliable estimates for smaller, under represented subgroups. However, their effectiveness often depends on assumptions that may not always hold true \cite{little1993post}. Despite these limitations, these persistent issues have prompted researchers to explore alternative data sources and cutting-edge technologies with the potential to improve the reliability and precision of public opinion research.

Among the technologies gaining significant traction in public opinion research is the generation of synthetic data by advanced language models, specifically Large Language Models (LLMs). Synthetic data refers to information created through computational processes that mimic real-world data patterns without direct observation. LLMs, which are trained on vast and diverse text datasets, have been proposed as a novel method for producing synthetic public opinion data \cite{argyle2023out}. 
These sophisticated models identify complex statistical relationships between demographic variables and the language used in political contexts. Beyond simple word associations, LLMs capture higher-order interactions during their training phase, optimizing the likelihood of sequences of words or phrases based on their contextual usage. This capability allows them to extrapolate from observed patterns, generating survey responses that aim to represent the political perspectives of various demographic groups. The generation process typically involves guiding the model with specific parameters that outline the characteristics of a hypothetical respondent, such as their demographic background or political stance. Conditioning the LLM in this way can produce answers to survey questions that strive to reflect these specified traits.

The ability of LLMs to produce synthetic data on a large scale has sparked interest among
social scientists as a possible way to overcome the difficulties in gathering representative survey data \cite{argyle2023out, qu2024performance}. However, there are ongoing concerns about the accuracy of this synthetic data, with skepticism regarding whether LLMs truly capture real-world public opinions. As a result, researchers are creating methods to evaluate and measure the quality associated with this process.

This paper systematically describes the different approaches that can customize LLMs for tasks such as synthesizing public opinions. 
We evaluated the pros and cons of these approaches. In addition, we present a novel technique called role creation based on knowledge injection to LLMs for simulating population traits, a type of in-context learning. Our experimental results show that this new approach can significantly improve the accuracy of LLM-based public opinion polling. In addition, we discuss the impacts of LLM-simulated public opinion polling, challenges, and future research directions.  

Specifically, we address the following research questions:

\begin{itemize}
    \item \textbf{RQ1:} Can survey opinion data be simulated using LLMs?
    \item \textbf{RQ2:} Can our framework inject specific role knowledge into LLMs as part of in-context learning to help them better mimic human responses?
    \item \textbf{RQ3:} If \textbf{RQ2} is satisfied, can this approach be generalized and made model agnostic?
\end{itemize}
\section{Background and Motivation}

\subsection{The Promises of Synthetic Public Opinions with LLMs}

Social scientists recognize the potential of LLMs for generating synthetic samples, primarily because these models can produce data without the logistical challenges associated with traditional methods. Unlike human respondents, LLMs can manage longer surveys while maintaining data quality, helping to reduce respondent fatigue and loss of focus \cite{bail2024can, messeri2024artificial}.

LLMs demonstrate human-like traits when simulating human
behavior and psychological processes. For example, some models have shown they can mimic human moral decisions and
behavioral patterns \cite{dillion2023can}. This alignment with human ethical decision-making processes is evident in
their capacity to predict and reflect actual moral decisions \cite{dillion2023can}. When the morally correct
action is obvious, LLMs usually opt for sensible answers; however, in uncertain circumstances, they---like humans---show doubt
\cite{scherrer2024evaluating}. Moreover, LLMs can predict various social behaviors such as trust, cooperation, and
competitive tendencies \cite{leng2023do, xie2024can, zhao2024competeai}. Additionally, LLMs can capture public perceptions of personality traits among notable individuals, showcasing their flexibility and
precision in mimicking human behaviors \cite{cao2024large}.


LLMs also exhibit potential for advancing public opinion research by offering novel ways to simulate political behaviors and preferences. They can evaluate the positions of politicians on key policy issues \cite{wu2023large} and gauge public views on contentious topics like climate change \cite{lee2024can}. These models can also serve as practical tools for estimating voter choices \cite{qi2024representation}. Furthermore, generative agents have been shown to accurately replicate participants' responses on the General Social Survey, matching how participants would answer their own questions two weeks later, including on topics such as political party affiliation and ideology \cite{park2024generative}. The ability of LLMs to generate synthetic samples suggests their potential value in estimating public opinion, particularly in contexts where traditional data collection methods are constrained, such as in non-democratic regimes (though see \cite{qi2024representation}). They might even be capable of predicting public reactions to future political events \cite{wang2024survey}.

The ability of LLMs to enhance public opinion research extends beyond generating synthetic data. These models can also play a supportive role in various research stages. For example, LLMs can pre-test new survey questions and assist in developing item scales \cite{bail2024can}. They can serve as substitutes for human respondents who drop out of longitudinal studies, thus helping to maintain sample integrity. Furthermore, LLMs can annotate open-ended data collected from human or synthetic samples with minimal supervision, streamlining the data analysis process \cite{ziems2024can}. While social scientists are optimistic about the potential of LLMs to revolutionize public opinion research, significant challenges remain in ensuring that the synthetic data generated accurately reflects human public opinion.

\subsection{Approaches to Customize LLMs for Public Opinion Research} 

LLM customization methods can be broadly classified into pre-training and post-training approaches. 

\begin{figure}
\includegraphics[width=\linewidth]{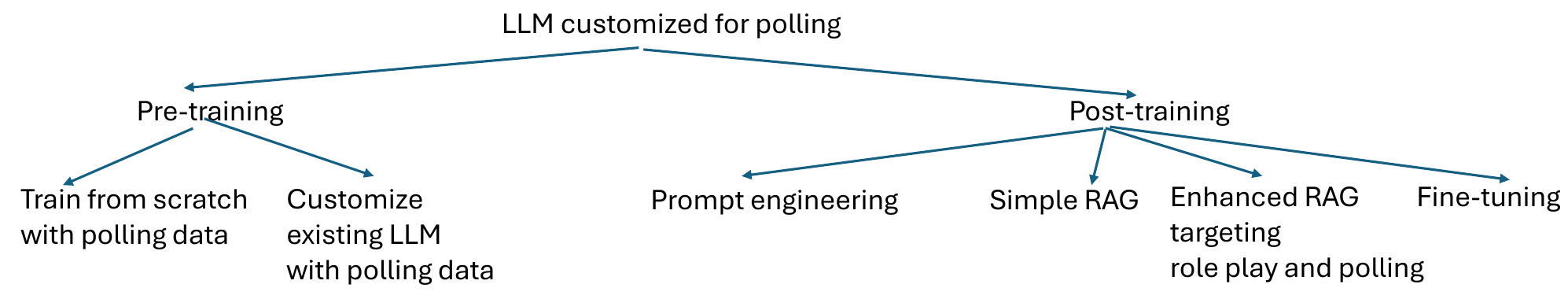}
\caption{Taxonomy of LLM task customization approaches.}
\label{fig:taxonomy}
\end{figure}

\subsubsection{Pre-training}
General-purpose LLMs are typically trained with open web data and can perform a wide range of tasks, such as responding to a survey. In the case of pre-training, with domain-specific datasets such as public opinion polling, an LLM can be trained from scratch for specific polling-related tasks such as role-playing and answering polling questions. LLMs trained in such a way can outperform general-purpose LLMs. The downside of this approach is the high cost and the time required to train a new LLM from the ground up. Recent advances in training, such as DeepSeek ~\cite{deepseekai2025deepseekv3technicalreport}, suggest a potential path for reducing training costs.  A more feasible approach for pre-training is to enhance existing general LLMs by further pre-training them with domain-specific data. Continuous pre-training involves taking pre-trained general-purpose LLMs and further training them on a new task or refining their ability to understand and perform within specific knowledge domains, such as public opinion polling and political attitude surveys.

\begin{table*}[ht]
\begin{center}
\caption{Comparison of different domain customization approaches.}
\begin{tabular}{llllll}
& \textbf{pre-training} & \textbf{prompt engineering} & \textbf{RAG}  & \textbf{Enhancements with role profiles} & \textbf{Fine tuning}      \\
Special knowledge & \Checkmark   &     & \Checkmark & \Checkmark & \Checkmark \\
Improved quality  & \Checkmark   & limited  & \Checkmark & \Checkmark, \Checkmark & \Checkmark \\
Model change     & \Checkmark, \Checkmark & no  & no  & no  & \Checkmark \\
Cost       & extremely high     & very low     & low       & low     & high         \\
Role customization   & \Checkmark       & limited    & \Checkmark & \Checkmark, \Checkmark & \Checkmark \\
Expertise required  & high   & low   & medium   & medium   & medium
\end{tabular}
\end{center}
\end{table*}

\subsubsection{Post-training}
In post-training approaches, a pre-trained LLM can be further refined with fine-tuning. Fine-tuning can tailor pre-trained models to the specific nuances of a task. Such specialization can significantly enhance the LLM's effectiveness in that particular task compared to a general-purpose pre-trained model. Like pre-training, fine-tuning incurs higher costs, requires AI expertise, and takes time. This method contrasts with other post-training approaches where the model weights are not changed. 

The most straightforward post-training approach is prompt engineering, which includes zero-shot and few-shot learning.  In the case of zero-shot learning, a user prompts an LLM without any examples, attempting to take advantage of the reasoning patterns it has gleaned in a general-purpose LLM. In zero-shot learning, a user provides an LLM with a prompt without any examples, aiming to leverage the reasoning patterns the model has acquired during its general-purpose training. Prompt engineering can enhance the performance of a pre-trained model; however, this improvement is often limited.  

Another line of post-training optimization is knowledge injection ~\cite{lauscher2020commonsenseworldknowledge,Chen_2022}. By incorporating knowledge bases such as political affiliation, polling results, political ideology, and demographics, it is possible to teach a pre-trained model about the specific domain and tasks. Fine-tuning is one way to add knowledge to a pre-trained model. With fine-tuning, the model builder continues the model training process and adapts it using task-specific data. By exposing the model to a specific knowledge domain, its weights can be adapted for the targeted applications. As a result,  its performance in particular tasks, such as polling simulation, can be more relevant to the specialized domains.

Another approach to improve a model's knowledge base is using in-context learning, such as retrieval augmented generation (RAG) ~\cite{fan2024surveyragmeetingllms, Lewis2020RetrievalAugmentedGF}. RAG utilizes information retrieval techniques to allow general-purpose LLMs to access relevant data from a knowledge source, often stored in vector databases, and integrate it into the generated text ~\cite{jing2024largelanguagemodelsmeet}. Post-training knowledge injection can address limitations for many knowledge-intensive tasks ~\cite{ovadia2024finetuningretrievalcomparingknowledge}. However, general post-training knowledge injections, such as existing RAG approaches, are not explicitly designed for role-play and opinion polling tasks. 

A promising direction is RAG enhanced with role profiles. RAG---augmented with role profiles---can simulate the political opinions of specific population groups more accurately than simple RAG. We introduce this approach to the literature and present experimental results to demonstrate its advantages for polling tasks.

\begin{figure}
\includegraphics[width=\linewidth]{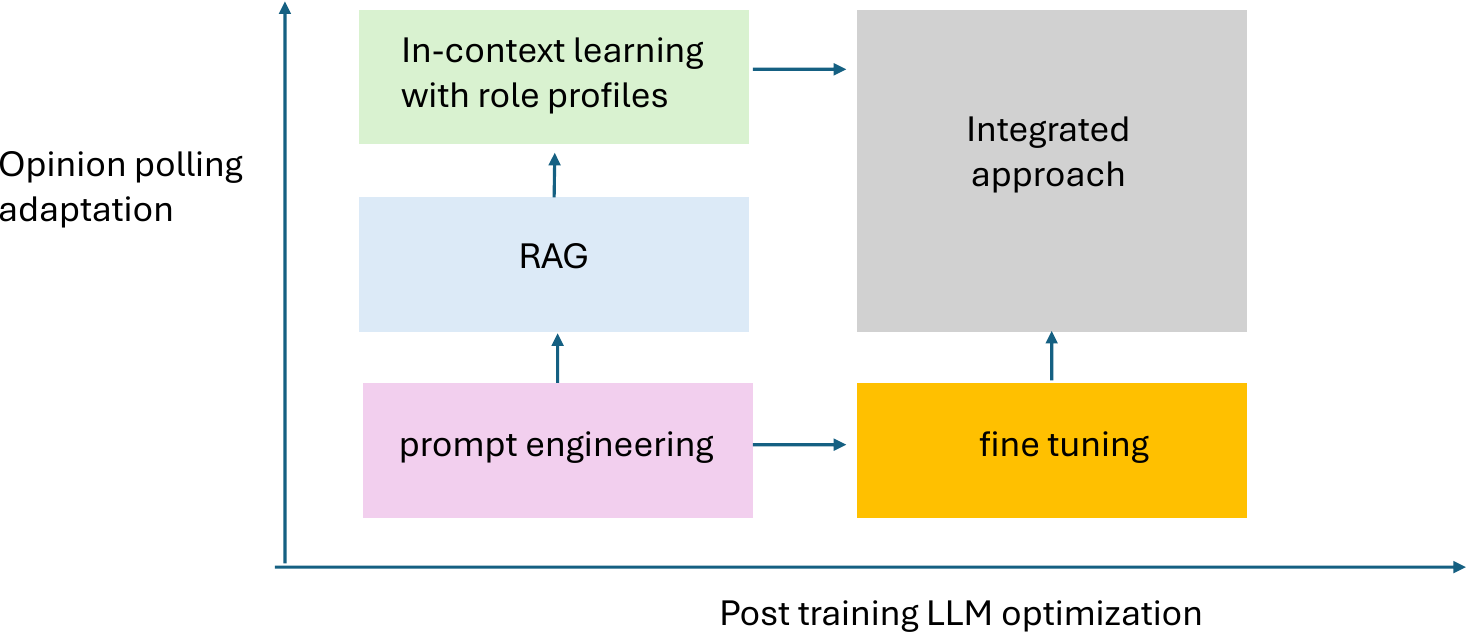}
\caption{Adapting LLMs to synthesizing public opinion tasks.}
\label{fig:rag}
\end{figure}

\section{Simulating Voter Preferences}

Several studies investigate the feasibility and effectiveness of using LLMs to simulate survey responses. A key study by \cite{argyle2023out} finds that LLMs can provide reasonably accurate simulations of group-level responses in behavioral science and economics experiments, as well as for political surveys. \cite{horton2023large} and \cite{aher2023using} further corroborate these findings, demonstrating the potential of LLMs to capture human-like response patterns in various survey settings.

One crucial aspect of simulating human responses is to capture the diversity of opinions across different cultural backgrounds. If trained effectively, LLMs have the potential to generate culturally nuanced responses, facilitate cross-cultural research, and provide insight into how cultural factors influence survey responses. Furthermore, LLMs can simulate such responses with varying levels of confidence. This capability allows for a more realistic simulation of human-like response patterns, as survey respondents often express their opinions with differing degrees of certainty.

However, existing research primarily focuses on using LLMs ``out of the box'' with prompting strategies. While prompt engineering can be practical, it may not fully capture the nuances of individual differences and personal traits that influence survey responses. This paper argues that by dynamically creating specific personal preferences in LLMs, we can achieve a more realistic and nuanced simulation of survey feedback.

\subsection{LLMs as Simulated Opinion Sources}

At their core, LLMs function as advanced statistical language models trained on vast datasets of textual information. These models predict the probability distribution of the next token (word or character) within a sequence based on preceding tokens. Mathematically, this process can be expressed as:

\[
p(x_n | x_1, \ldots, x_{n-1}),
\]
where \(x_i\) represents a token from a predefined vocabulary. This capability extends beyond mere
memorization, as it leverages the model's ability to capture intricate statistical patterns within training data, enabling novel and contextually appropriate text generation.

A key feature that renders LLMs particularly adept at simulating diverse opinions is their reliance on conditioning. Before generating text, the model receives an initial input or context that significantly influences subsequent output. 
This conditioning context, represented by tokens \(x_1, \ldots, x_{n-1}\), plays a crucial role in guiding the model's response. By strategically modifying this context, we can exert substantial control over the direction of text generation. 
For instance, providing a context that outlines specific demographic traits or political orientations can alter the probability distribution of subsequent tokens, prompting responses aligned with specified characteristics.

\begin{table*}[htbp] 
\centering 
\footnotesize 
\setlength{\tabcolsep}{3pt} 
\renewcommand{\arraystretch}{0.9} 
\caption{Different Role Creation based on Attributes}
\label{tab:roles}
\begin{tabularx}{\linewidth}{|>{\RaggedRight\hsize=0.2\hsize}X| >{\RaggedRight\hsize=0.25\hsize}X| >{\RaggedRight\hsize=0.55\hsize}X|} 
\hline
\textbf{HEXACO} & \textbf{Political} & \textbf{Concise Prompt (Respond as...)} \\ \hline
\textbf{Dimension} & \textbf{Leaning} &  \\ \hline
\multirow{3}{*}{\textbf{H}} & Conservative & \texttt{...high honesty/humility, conservative values. Integrity, tradition, honest governance.} \\ \cline{2-3}
 & Liberal & \texttt{...high honesty/humility, liberal justice. Equality, fairness, systemic equity.} \\ \cline{2-3}
 & Populist & \texttt{...cynical of power. Low H elites, skeptical, 'common person' vs 'establishment'.} \\ \hline
\multirow{3}{*}{\textbf{E}} & Conservative & \texttt{...low emotionality, national security. Calm, rational, strong defense, measured concern.} \\ \cline{2-3}
 & Liberal & \texttt{...high emotionality, social empathy. Concerned, empathetic, vulnerable, compassionate solutions.} \\ \cline{2-3}
 & Populist & \texttt{...emotional appeals, common frustrations. High E grievances, overlooked, wronged by elites.} \\ \hline
\multirow{3}{*}{\textbf{X}} & Conservative & \texttt{...introverted, measured action. Deliberate, reserved, cautious, behind-scenes influence.} \\ \cline{2-3}
 & Liberal & \texttt{...extraverted, public engagement. Lively, engaging, activist, public discourse, collective action.} \\ \cline{2-3}
 & Populist & \texttt{...extraverted, rally base. Energetic, direct, 'common person', bypass 'establishment'.} \\ \hline
\multirow{3}{*}{\textbf{A}} & Conservative & \texttt{...low agreeableness, firm stance. Direct, less consensus, strong convictions, principled.} \\ \cline{2-3}
 & Liberal & \texttt{...high agreeableness, consensus. Cooperative, polite, common ground, compromise, harmony.} \\ \cline{2-3}
 & Populist & \texttt{...low agreeableness vs. elites. Combative, critical, 'people's will', conflict if needed.} \\ \hline
\multirow{3}{*}{\textbf{C}} & Conservative & \texttt{...high conscientiousness, fiscal responsibility. Organized, rules, disciplined, efficient, responsible gov.} \\ \cline{2-3}
 & Liberal & \texttt{...low conscientiousness flexible, urgent needs. Flexible, responsive, immediate problems, adaptable policy.} \\ \cline{2-3}
 & Populist & \texttt{...low conscientiousness anti-bureaucracy. Disregard 'red tape', direct action, swift results.} \\ \hline
\multirow{3}{*}{\textbf{O}} & Conservative & \texttt{...low openness, tradition. Conventional, historical precedent, cautious change, proven methods.} \\ \cline{2-3}
 & Liberal & \texttt{...high openness, progress. Creative, forward-thinking, innovative, social progress, rethink systems.} \\ \cline{2-3}
 & Populist & \texttt{...high openness disruptive style. Reject 'elitist' norms, unconventional style, disrupt status quo.} \\ \hline
\end{tabularx}%

\end{table*}

\begin{figure}
    \centering
    \includegraphics[width=1\linewidth]{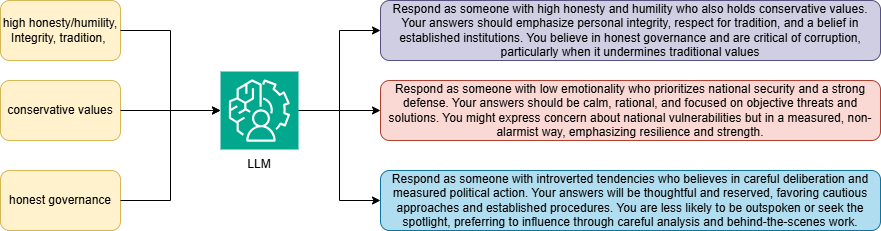}
    \caption{Role generation from attributes.}
    \label{fig:text2role}
\end{figure}

\tablename \ref{tab:roles} provides examples of how the profile creation from the attribute pool works. Figure \ref{fig:text2role} shows how the LLM takes these attributes and uses them---alongside political leanings---to create the \textit{roles} to be used for opinion generation.

\begin{figure}
    \centering
    \includegraphics[width=1\linewidth]{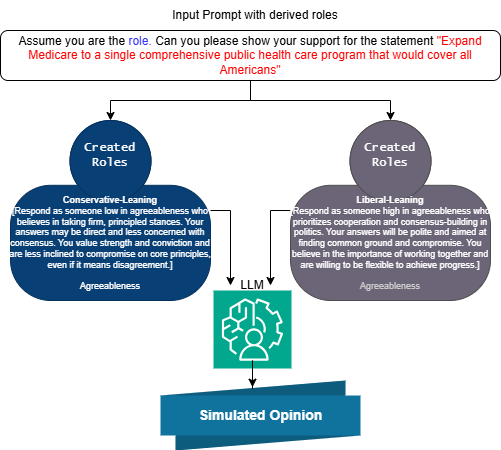}
    \caption{Opinion generation using roles.}
    \label{fig:role2e}
\end{figure}

Once we have the roles created, as shown in Figure \ref{fig:text2role}, we save those in a vector database to be used for a dynamic RAG for querying our LLM before sending the query to each user. Once a question is asked, the RAG outputs a section of the relevant profile that matches with the few-shot prompts passed through the query. Based on this retrieved RAG response, we query the LLM for an opinion/answer to the queried question, asking the LLM to role-play according to the retrieved profile. This retrieved profile is dynamic and can be potentially infinite based on the variation of the few shot prompts. We illustrate the full architecture in Figure \ref{fig:role2e}.

\section{Experiments and Evaluation}\label{sec:result}

We evaluate our framework and survey results by comparing them with existing human evaluation results. For our evaluation, we ask the same questions as our dataset and compare the responses' similarity with human-given responses. This dataset has not been used for training, nor were any few-shot examples used for our framework or the pre-trained LLMs with which we compare.

\subsection{Data}
Our study centers on 30 issue-related questions from the 2021 Cooperative Election Study (CES), previously known as the Cooperative Congressional Election Study (CCES). We chose to focus on issue attitudes because they are the most frequent and influential targets of political polls \cite{morris2022strength}. Extensive observational and experimental evidence demonstrates that these issue polls significantly influence political decision-making \cite{burstein2003impact, butler2011can, wlezien2016public, morris2022strength}. Consequently, these polling responses are those most likely to be sought from synthetic respondents. Among available data sources, we selected the CCES due to its substantial respondent base (exceeding 17,000) and consistent coverage of a broad range of issues. We focused on 2021 as the most recent year for which we could reasonably assume that all the LLMs under consideration had been trained on relevant data.\footnote{It is crucial to acknowledge that the training data for LLMs typically lags behind the LLM's publication date by at least a couple of years. At this time, real-time updating of LLM weights is impractical due to resource constraints.}

\subsection{Experimental Setup}

We ran all our experiments on a machine running on Ubuntu 24.04 with RTX 4090 as our inference provider. llama.cpp was the backend with Vulkan API support to run and generate the experiments. We intentionally generated every response in a fresh state without memory of the previous interactions. All the experiments were done, keeping the model temperature at 0.

\subsection{Result Adherence}

We begin by assessing the accuracy of the LLM's responses in comparison to human responses, focusing specifically on how well LLMs replicate human responses. A key distinction between our study and previous ones is that we not only examine aggregate results but also analyze individual opinion levels to determine how closely they match human responses. We analyze both the respondent-level accuracy as well as topic-level accuracy. We compare the responses of various LLMs and humans with those generated by our framework to see if our methodology produces results that closely align with human responses.

\subsubsection{Accuracy for Individual Responses}

For individual responses, we begin by examining the issues where the LLM's responses aligned with human responses, measured by percentage.

\begin{table}[]
\caption{Simulated Result Adherence}
\label{tab:rsult}
\resizebox{\columnwidth}{!}{%
\begin{tabular}{|l|l|l|}
\hline
Model                     & Parameters & Adherence On Questions(\%) \\ \hline
phi3                      & 14b        & 72.7                       \\ \hline
deepseek-r1               & 32b        & 79.8                       \\ \hline
gemma2                    & 9b         & 70.2                       \\ \hline
gemma2                    & 27b        & 71.6                       \\ \hline
llama3.2                  & 3b         & 67.6                       \\ \hline
llama3.3                  & 70b        & 73.1                       \\ \hline
Role Creation + gemma2    & 27b        & 77.9                       \\ \hline
Role Creation + gemma2    & 9b         & 73.3                       \\ \hline
Role Creation + llama 3.3 & 70b        & 84.1                       \\ \hline
\end{tabular}%
}

\end{table}


Table \ref{tab:rsult} shows our adherence to different topics with the questions of the CCES questionnaire. We used a few-shot prompting for the different pre-trained LLMs before getting the results, But we used the template shown in Figure \ref{fig:role2e} to generate opinions for the role generation.

\section{Discussion}

Based on our empirical experiments shown in Section \ref{sec:result}, we can observe certain specific characteristics of the survey response for the models. While we targeted particular questions, we have achieved to show the adherence trends in Table \ref{tab:rsult}. 

If we circle back to our original queries:

\begin{tcolorbox}[
  colback=green!15, 
  colframe=green!40, 
  title=RQ1: Can survey opinion data be simulated using LLMs?,
  coltitle=black, 
  fonttitle=\bfseries,
  boxrule=0.75mm, 
  rounded corners, 
  left=1mm, 
  right=1mm, 
  top=1mm, 
  bottom=1mm 
]
\textbf{Yes.} The results in Table \ref{tab:rsult}, clearly indicate that opinions can be simulated by LLMs. However, the method by which we can simulate these opinions does vary the achieved adherence performance. For all the responses, we had to provide the voter demography and profile information either by a few-shot prompt or through our framework's role creation.
\end{tcolorbox}

While we utilize open-weight LLMs with few shot prompting, our framework uses dynamic role creation based on personality attributes as depicted in Table\ref{tab:roles} and voter preferences. This method allows us to dynamically generate multiple roles with granular preferences attached to them, as shown in Figure \ref{fig:text2role}. For our question, we then use a RAG system to select these roles and---as the LLM---to give opinions on the survey questions following the flow shown in Figure \ref{fig:role2e}. The text highlighted in blue are the dynamically generated roles from Figure \ref{fig:text2role} retrieved from RAG using the question and then sent to the LLM to get a simulated response. 

These results bring us to our second research question:

\begin{tcolorbox}[
  colback=green!15, 
  colframe=green!40, 
  title=RQ2: Can response adherence be increased for LLMs to mimic humans?,
  coltitle=black, 
  fonttitle=\bfseries,
  boxrule=0.75mm, 
  rounded corners, 
  left=1mm, 
  right=1mm, 
  top=1mm, 
  bottom=1mm 
]
\textbf{Yes.} Our empirical results in Section \ref{sec:result} show that using our framework of dynamically generating roles and then pairing them up with existing pre-trained LLMs increases their adherence to human responses.

\hspace{0.5cm} We test our framework with three different LLMs from two different LLM families, and they all show improvement from their base models, which were prompted with few-shot prompts. One noticeable insight we glean is that the bigger models show bigger gains with the same technique than the smaller models. This finding suggests that the ability to understand nuanced instructions likely played a role in how closely the generated opinions aligned with human responses.
\end{tcolorbox}

This result, however, raises an interesting question. Can the LLMs predict human responses, or do they already have these associations as part of their pretraining data? Judging from how all the LLMs crossed 50\% adherence with human responses with just a few-shot prompting, we hypothesize the preference associations are already present in the LLM's pretraining data. Making them roleplay with targeted \textit{roles} seems to further encourage more opinionated responses.

That finding brings us to our last research question:

\begin{tcolorbox}[
  colback=green!15, 
  colframe=green!40, 
  title=RQ3: Can we generalize this framework and make it model agnostic?,
  coltitle=black, 
  fonttitle=\bfseries,
  boxrule=0.75mm, 
  rounded corners, 
  left=1mm, 
  right=1mm, 
  top=1mm, 
  bottom=1mm 
]
Our empirical and experimental results show improvements across three different models. However, a closer examination reveals inconsistencies in the actual performance gains, highlighting a weakness in our framework. Since we primarily rely on the LLMs’ ability to understand roles and elicit opinionated responses, their performance suffers when they fail to grasp complex instructions or nuances.
\end{tcolorbox}

These observations lead us to consider whether we can embed models' preferences through fine-tuning and trigger-based generation. We leave this as an open research question for future work, along with another question regarding the role of non-English languages in generating similar opinions.
\section{Impacts}

Using LLMs to synthesize public opinions has the potential to transform the democratic process on a global scale. For decades, researchers have gathered public opinion data through labor-intensive, costly, and time-consuming methods such as in-person interviews, phone calls, and survey mailings. Additionally, gathering accurate public opinions can be challenging in regions with low economic development, and in some countries, local governments actively control or restrict public opinion surveys. The LLM-based approach offers a promising solution to these challenges by making the process cheaper, faster, and more accurate. For example, it can refine survey questions using simulated polls before finalizing them, reducing bias and improving question quality. Furthermore, LLM role-playing in public opinion simulation opens a new frontier where campaigns can model citizens' reactions to different candidate messages. This pre-testing allows campaigns to experiment with various strategies and messages before implementation. Crucially, this method also enables the simulation of responses from populations whose opinions might otherwise be marginalized or silenced.

\section{Challenges and research directions}

\subsection{Technology Limitations}
The existing literature identifies potential limitations of using LLMs to generate synthetic samples for public opinion research. These include the risk of training data memorization, where models might reproduce specific details instead of developing new inferences, and sensitivity to prompt formulations, which can lead to inconsistent or biased outputs. Variations across different LLMs can also compromise the reliability of generated samples, and the models' tendency to generalize may introduce distortions.

LLMs are sensitive to prompts' precise wording and structure, which can substantially influence their outputs. When applying linguistic rules and world knowledge, LLMs can be influenced by specific examples and phrasing, leading to response variations based on subtle changes in prompt formulation \cite{chang2024language}. This sensitivity becomes less significant when employing our technique with role creation. A thorough investigation into how our role creation technique enhances RAG and in-context learning to mitigate the sensitivity of LLMs to prompt variations would provide valuable insights and is a promising direction for future research.

Another concern in using LLMs to generate synthetic samples for public opinion research is the inconsistency across different models. Research demonstrates that different LLMs can exhibit varying traits and performance levels, potentially leading to output discrepancies. To mitigate this issue, researchers either need a deep understanding of the relative strengths and weaknesses of different models, or they must be able to identify high-quality models to include in an ensemble model, which could mitigate inconsistencies.

\subsection{Localization and Less Represented Languages} A significant challenge with large language models, especially in political polling tasks and when working with languages other than English, is the notable gap in available training data. This discrepancy results in poorer performance for LLMs in less commonly spoken languages due to limited data, imbalances, biases, and cultural nuances. The practical use of LLMs for polling in non-English languages and non-Western cultures presents significant challenges, making it an important area for future research.

\subsection{Data Quality} Knowledge injection and creation of role profiles depend on data quality. Public opinions and policy preferences shift over time. The quality of the data used in different phases of knowledge injection and fine-tuning, as well as the timeliness of the data, can affect the accuracy of polling tasks.

\subsection{Continuous Updates and Learning} For public opinion polling, the RAG database and role profiles must be regularly updated to capture shifts in public opinion, demographic changes, and reactions to major political events that can influence views on policies and political decisions. This requires ongoing updates to the knowledge base, which increases both labor and financial costs. Finding ways to reduce these costs without compromising the accuracy of the LLM in polling tasks presents a significant research challenge.

\subsection{Compliance} With the wider adoption of LLMs for polling, compliance could become a challenge. In some countries, these capabilities may be exploited or manipulated to spread false information and deceive the public. Ensuring public trust in LLM-based polling could be a significant issue in the future.

\section{Conclusion}
This research demonstrates the potential of LLMs to create synthetic public opinion data and contributes a novel method that goes beyond a few short prompts to generate more precise responses. Using a RAG method with user roles, the proposed role-creation framework significantly enhances the accuracy of simulated opinions, mitigating issues identified with standard prompting or basic RAG implementation. The results show improved answer adherence between model opinion and existing human dataset. The potential impact of this work on improving the cost and time for opinion collection, along with democratic processes---particularly in under-resourced or restrictive environments---is substantial, opening new avenues for eDemocracy.

\section*{Acknowledgment}

We would like to express our sincere gratitude to the Google Developer Expert program and the Google ML team for their invaluable support in this research, providing GCP credits that enabled the successful execution of this work.

\bibliographystyle{plainnat}
\bibliography{ref}



\end{document}